\documentclass[10pt,twocolumn,letterpaper]{article}

\usepackage{iccv}
\usepackage{times}
\usepackage{epsfig}
\usepackage{graphicx}
\usepackage{amsmath}
\usepackage{amssymb}
\usepackage{comment}
\usepackage{booktabs}

\usepackage{multirow}

\newcommand\blfootnote[1]{%
  \begingroup
  \renewcommand\thefootnote{}\footnote{#1}%
  \addtocounter{footnote}{-1}%
  \endgroup
}

\usepackage[breaklinks=true,bookmarks=false]{hyperref}

\iccvfinalcopy 

\def\iccvPaperID{3091} 

\ificcvfinal\fi

\begin{document}

\title{Move to See Better: Self-Improving Embodied Object Detection}

\author{
  Zhaoyuan~Fang$^{\dagger}$ \\
  Carnegie Mellon University \\
  \texttt{zhaoyuaf@andrew.cmu.edu} \\
  \and
  Ayush~Jain$^{\dagger}$ \\
  Carnegie Mellon University\\
  \texttt{ayushj2@andrew.cmu.edu} \\
  \and
  Gabriel~Sarch$^{\dagger}$ \\
  Carnegie Mellon University \\
  \texttt{gsarch@andrew.cmu.edu} \\
  \and
  Adam~W.~Harley \\
  Carnegie Mellon University\\
  \texttt{aharley@cmu.edu} \\
  \and
  Katerina~Fragkiadaki \\
  Carnegie Mellon University\\
  \texttt{katef@cs.cmu.edu} \\
}
\maketitle
\blfootnote{$^{\dagger}$ The authors contributed equally. The order is sorted alphabetically by last names.}

\maketitle
\ificcvfinal\thispagestyle{empty}\fi

\begin{abstract}
   Passive methods for object detection and segmentation treat images of the same scene as individual samples, and do not exploit object permanence across multiple views. Generalization to novel or difficult viewpoints thus requires additional training with lots of annotations. In contrast, humans often recognize objects by simply moving around, to get more informative viewpoints. In this paper, we propose a method for improving object detection in testing environments, assuming nothing but an embodied agent with a pre-trained 2D object detector. Our agent collects multi-view data, generates 2D and 3D pseudo-labels, and fine-tunes its detector in a self-supervised manner. Experiments on both indoor and outdoor datasets show that (1) our method obtains high quality 2D and 3D pseudo-labels from multi-view RGB-D data; (2) fine-tuning with these pseudo-labels improves the 2D detector significantly in the test environment; (3) training a 3D detector with our pseudo-labels outperforms a prior self-supervised method by a large margin; (4) given weak supervision, our method can generate better pseudo-labels for novel objects. 
\end{abstract}
\section{Introduction}
\begin{figure}[tb]
\begin{center}
\includegraphics[width=0.98\linewidth]{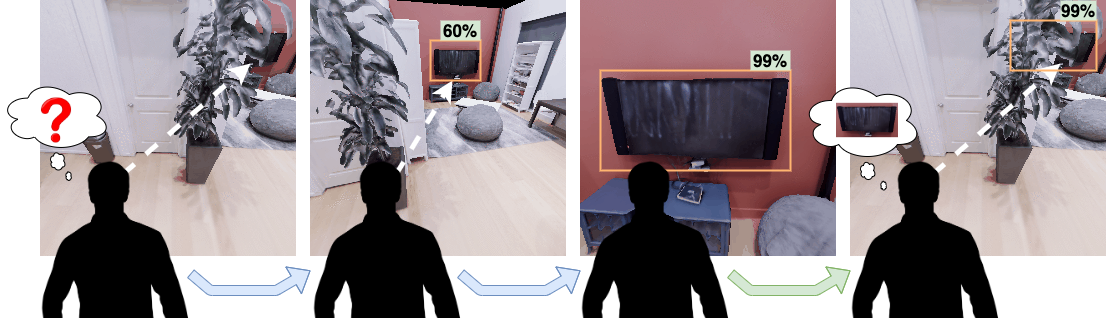}
\end{center}
   \caption{\textbf{Improving object recognition by moving.} An agent is viewing an object from an occluded, unfamiliar viewpoint. By moving to less occluded, more familiar viewpoints of the object (blue arrow), the agent can use the familiar viewpoints to self-supervise the previously unfamiliar viewpoints (green arrow).
   }
\label{fig:teaser}
\end{figure}

For tasks that require high-level reasoning, intelligent systems must be able to recognize objects despite partial occlusions or uncommon poses. 
Humans and other mammals actively move their eyes, head, and body to obtain less occluded and more familiar viewpoints of the objects of interest \cite{edelman1992orientation, tarr1995rotating}. They then use familiar viewpoints to inform viewpoints they are less confident about. 
For example, imagine a scenario where one is attempting to recognize an occluded object from an unfamiliar viewpoint, such as the TV in Figure~\ref{fig:teaser}. An intelligent agent can increase its accuracy in this task simply by moving to a less occluded and more familiar viewpoint, and then mapping these confident beliefs of the object back to the unfamiliar views.




Significant improvements have been made in the accuracy and reliability of 2D
and 3D
visual recognition systems~\cite{liu2020detsurvey}. 
However, recent works~\cite{Barbu2019ObjectNet, purushwalkam2020demystifying} show that current detectors are less likely to recognize an object correctly under occlusions and uncommon viewpoints. Advances in active visual learning~\cite{chaplot2020semantic, chaplot2020neuralslam, Yang2019ICCVembodied} have focused on efficient data collection techniques, so that the detector adapts to new scenes and views after fine-tuning on the collected data. However, these approaches require either ground truth 3D segmentation of the environment or 2D human annotations of the images (in addition to a pre-trained detector). 

 
In this work, we demonstrate that by obtaining a diverse set of views of an object and propagating confident detections across viewpoints, we can increase detector performance without any additional annotations. We propose a method for an embodied agent to improve its 2D and 3D object detection in test environments, assuming only a pre-trained 2D object detector, a depth sensor, and approximate egomotion. We show robustness to realistic actuation noise as well. Active agents typically encounter novel objects in their environment. Our weakly supervised method can generate pseudo-labels for them that can be used to successfully train object detectors with significantly less human annotations. 
Our method provides a robust unsupervised way to obtain pseudo-labels for 3D detection and segmentation. The detector improvement procedure has three phases: \textit{(1) Data collection}: randomly move in the environment to collect observations and occasional high-confidence detections, then plan paths to collect diverse posed RGB-D images of the detected objects; \textit{(2) 3D segmentation}: segment the detected objects in 3D using aggregated RGB-D images, and then re-project that segmentation to form 2D pseudo-labels in all views; \textit{(3) Detector improvement}:  fine-tune the pre-trained detector on the pseudo-labels.  
 
 
Our key contribution is a 2D and 3D pseudo-label generation method, allowing object detectors and segmentors to be fine-tuned self-supervised. Our method assumes an embodied agent with a pre-trained 2D object detector, a depth sensor, and approximate knowledge of its egomotion. We show that fine-tuning with labels generated by our method significantly improves the performance of the pre-trained detector in challenging indoor and outdoor datasets. Adding weak supervision further increases performance. Moreover, we can learn object detectors for novel categories given weak supervision. Additionally, we show that our self-supervised 3D detector outperforms a state-of-the-art self-supervised 3D detection method by a large margin, while achieving performance comparable to a fully supervised model with the same architecture. 
We have made our code publicly available to facilitate future research \footnote{https://github.com/ayushjain1144/SeeingByMoving}.  

\section{Related Work}
\textbf{2D Object Recognition}\quad
Deep networks achieve good performance on 2D object detection~\cite{Ren2015FasterRCNN} and segmentation~\cite{He2017MaskRCNN}. Data augmentation techniques~\cite{Dwibedi_2017_cutpastelearn, Zoph_2020_aug} have also shown promising results to enhance training with scarce annotations. However, recent works~\cite{Barbu2019ObjectNet, purushwalkam2020demystifying} show that current detectors are less likely to correctly recognize an object from uncommon viewpoints, and augmentation methods fail to capture viewpoint invariances which are essential for 2D recognition systems. In this work, we aim to improve a pre-trained detector in \textit{new environments and viewpoints}, by taking advantage of an embodied agent's opportunity to move around and collect difficult views in the environment. 

\textbf{3D Object Recognition}\quad
3D object recognition has also been explored in various forms. Some methods quantize pointclouds into 3D voxel grids~\cite{ren18sbnet,Zhou2018VoxelNet}. PointNet~\cite{qi2016pointnet, qi2017pointnetplusplus, qi2017frustum} directly operates on unordered pointclouds for learning deep point set features applicable to object detection and segmentation. SPGN~\cite{wang2018sgpn} extends the direct consumption of pointcloud to instance segmentation. Later works~\cite{Li2018PointCNN, Wang2018DeepParametric} integrate convolution into pointclouds. All of these methods require 3D annotations. Armeni et al.~\cite{3dsg} uses a pre-trained detector to build a 3D scene graph in a semi-automatic way, and embeds its knowledge onto the 3D scene representation akin to a SLAM method. Our method instead embeds its knowledge into detectors, which allows it to work even under settings where the scenes change over time.  LDLS~\cite{Wang2019LDLS} performs 3D segmentation in a semi-supervised way, diffusing pre-trained detectors' predictions from RGB-D images onto the scene pointcloud. In this work, we show that with our pseudo-labels we can train a 3D object detector~\cite{qi2017frustum} 
that outperforms LDLS and achieves performance comparable to the same detector trained on ground truth labels. 
\begin{figure*}[!htb]
\begin{center}
\includegraphics[width=0.9\linewidth]{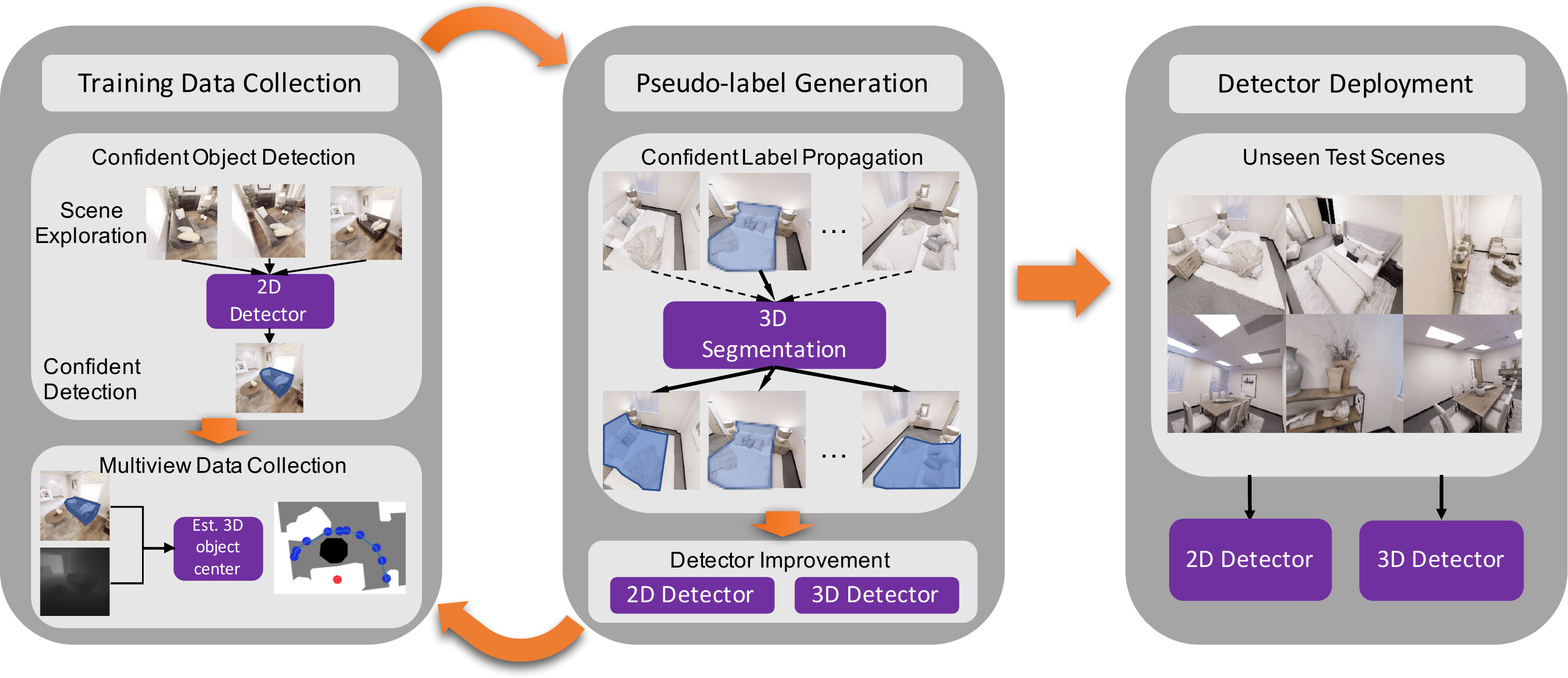}
\end{center}
   \caption{\textbf{Seeing by Moving (SbM).} We use confident detections of a pre-trained 2D object detector to guide self-supervised multi-view data collection and pseudo-label generation. Our 3D segmentation module segment the detected objects in 3D using aggregated RGB-D images. The 2D and 3D detectors fine-tuned on the pseudo-labels perform better on unseen test scenes.}
\label{fig:pipeline}
\end{figure*}

\textbf{Active Visual Learning}\quad 
The problem of active vision~\cite{Aloimonos1988Active, Ballard1991AnimateVision, settles2009active} presents an agent with a large unlabelled set of images and asks the agent to select a subset for labelling, which will provide the maximal amount of information about the full dataset~\cite{sener2018active}. Psychology research suggests active vision as a natural method used by humans to attend to relevant visual features~\cite{Bambach2018Toddler, edelman1992orientation, tarr1995rotating, yang2016active}. This has been applied to several fields including object detection~\cite{Jayaraman2019Endtoend, Johns2016PairwiseDO, Vijayanarasimhan2011LargeActive, Yang2018CoRLVisualCuriosity}, instance segmentation~\cite{pathakCVPRW18segByInt}, feature learning~\cite{Agrawal2015SeebyMoving}, and medical image analysis~\cite{Kuo2018CostSensitiveAL}. Chaplot \etal~\cite{chaplot2020semantic} explored a closely related setting, where a policy is trained to efficiently acquire data where detections are not multi-view consistent. 
In this work, we propose a self-supervised technique complementary to both directions, where we select ``easy" viewpoints according to the confidence of a pre-trained detector, and propagate information from these viewpoints to more challenging ones.

\textbf{Embodiment}\quad
Embodied agents can move and interact with their environment through a physical apparatus. 3D simulators have been an important part of modelling embodiment in a virtual setting. Many of the environments are photo-realistic reconstructions of indoor~\cite{replica19arxiv, ammirato2017dataset, xia2018gibson, chang2017matterport3d} and outdoor~\cite{Dosovitskiy17CARLA, geiger2013vision} scenes, and provide 3D ground truth labels for objects. These simulated environments have been used to study tasks such as visual navigation and exploration~\cite{chaplot2020neuralslam, gupta2017cognitive, fang2019scene}, visual question answering~\cite{das2018embodied}, tracking~\cite{Harley2020Tracking}, and object recognition~\cite{chaplot2020semantic, Yang2019ICCVembodied}. In our work, we use a simulated embodied agent to discover objects and collect diverse posed data for fine-tuning a detector. 

\section{Move to See Better}
We propose a method for an embodied agent to improve its 2D and 3D object detection in unseen environments, assuming only a pre-trained 2D object detector, a depth sensor, and self-provided egomotion information. Most previous methods that attempt to improve detection of embodied agents~\cite{chaplot2020neuralslam, chaplot2020semantic, Yang2019ICCVembodied} require either 2D or 3D human annotations after they have been collected by the embodied agent. Some of those methods train the movement of the agent to select specific viewpoints for later labelling~\cite{chaplot2020semantic, Yang2019ICCVembodied}. However, acquiring the annotations for the collected images still remains extremely expensive. 

We introduce a ``Seeing by Moving'' (SbM) framework, which removes the bottleneck of expensive human annotations, by driving the annotation process with the agent itself. 
An overview of our framework is shown in Figure~\ref{fig:pipeline}. In the data collection stage, we take advantage of the classifier head in a pre-trained object detector, which has high confidence when the object is viewed unoccluded in a common pose. The confidence values of the pre-trained detector serve as a cue to help us select good views of objects in the agent's environment. Although the confidence of a detector is not always well-calibrated with its accuracy, we are able to maintain high precision by setting a strict confidence threshold (see supp. for more details). In the pseudo-label generation stage, we propagate the high-confidence detections from ``easy" views to ``hard" views. 
We then fine-tune the 2D object detector using generated pseudo-labels. We demonstrate large improvements in both indoor and outdoor benchmarks. Additionally, we train a 3D detector from scratch using the pseudo-labels. 
Experimental results 
are discussed in Section~\ref{sec:exp}.

\subsection{Embodied Data Collection}
The aim of our data collection policy is to capture a diverse set of viewpoints of the objects present in the environment. Note that the datasets used for training deep object detection models most often capture objects from unoccluded and canonical viewpoints. In our data collection policy, we seek to obtain these unoccluded canonical viewpoints \textit{as well as} other viewpoints where a pre-trained detector would be less certain. Since we perform our experiments in simulation, it is possible to simply obtain a diverse set of viewpoints directly from the simulator, but we present a more general method here that could work in real-world scenarios as well. 

We consider an embodied agent equipped with a pre-trained object detector, a depth sensor, and approximate egomotion information. Data collection proceeds in object-centric \textit{episodes}. Episodes broadly have two stages: localizing a random object, then collecting $N$ views. 

To localize a random object, the agent naively explores the scene with a random policy and runs the detector on every frame. When the detector returns a sufficiently confident detection (determined by a threshold), we proceed to collect additional views of that object.

To collect views, the agent needs to navigate to positions at various viewing angles and distances from the object. We begin by estimating the 3D centroid of the object, using the estimated 2D object mask, the depth map, and the camera intrinsics. We then unproject (see section: \ref{sec:seg}) the depth map to construct a 3D occupancy map of the region, and use this map to sample a valid navigation location near the agent and within a distance from the object centroid, similar to Gupta \etal~\cite{gupta2017cognitive}. Given the occupancy map and goal location, we use a fast marching planner to reach the goal~\cite{sethian1996fast}. 
Once we reach the goal location, we use the object centroid and estimated pose to orient the viewing angle of the sensors so that the object is in view, and capture the sensor readings (i.e., the RGB-D image and the pose). 
We repeat this navigation and view-capture process until $N$ views have been obtained. At the end of an episode, we navigate away from the target object, and restart the random localization process. We collect 30 such episodes per environment.

To simulate actuation noise as observed in a real-world robot, we apply the noise model from Chaplot \etal~\cite{chaplot2020neuralslam}. This fits a separate Gaussian Mixture Model for each action (move forward, turn right, turn left) based on noise measurements from LoCoBot\footnote{http://locobot.org}. To register a cleaner scene pointcloud from the noisy measurements, we refine the camera pose by optimizing a cycle consistency objective based on flow from depth (see supp. for more details). We report experimental results both with and without this noise model in Section~\ref{sec:exp}. 

\begin{figure}[!htb]
\begin{center}
\includegraphics[width=\linewidth]{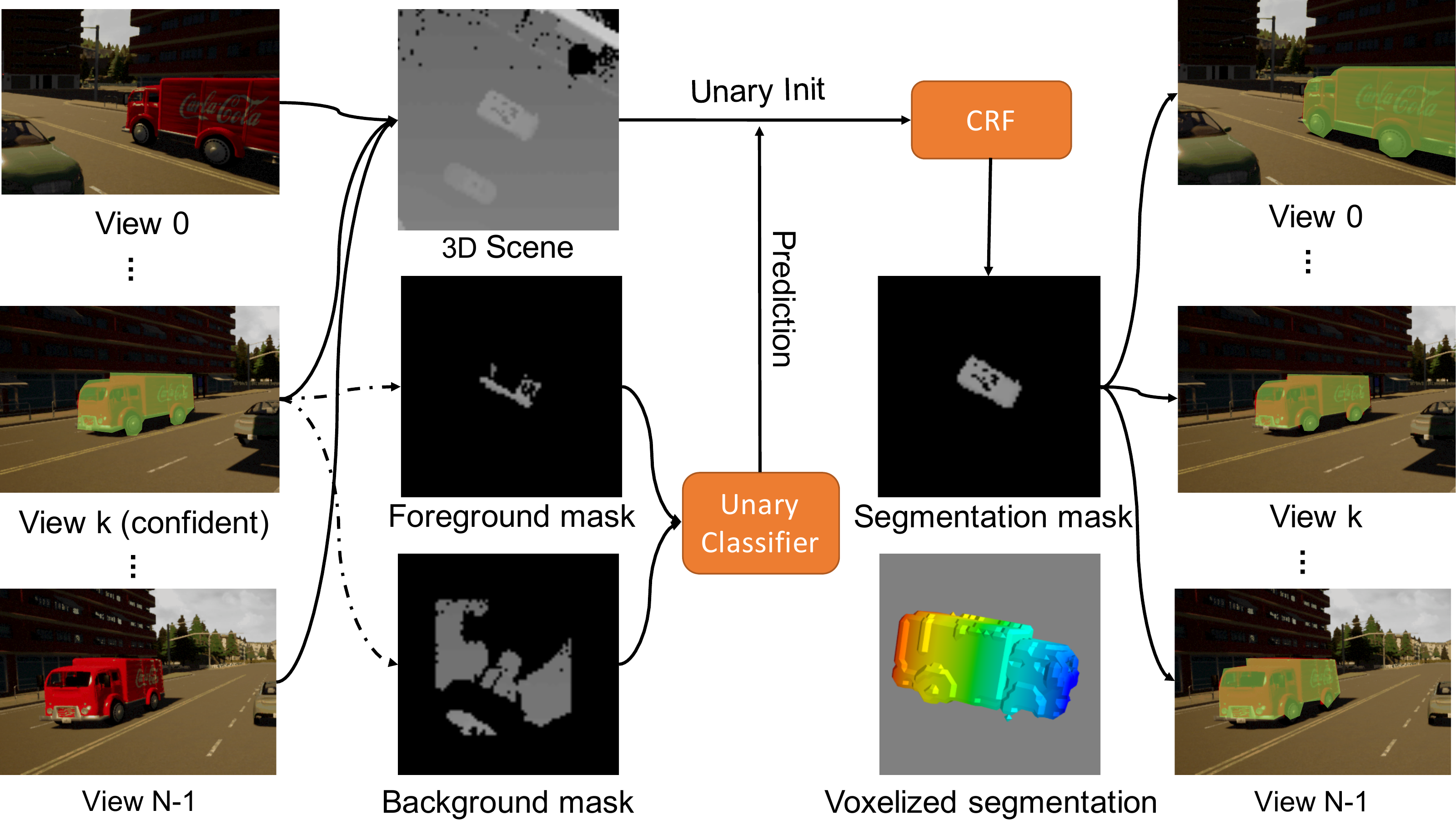}
\end{center}
\caption{\textbf{3D Segmentation.} All images are unprojected into 3D space using depth and pose, as well as the segmentation mask from confident view $k$. We sample foreground and background to train a unary classifier, whose outputs are used to initialize the unary potentials of the CRF model. The final 3D segmentation is then reprojected to all views to obtain pseudo-labels.}
\label{fig:model}
\end{figure}
\subsection{Multi-View Single-Object Segmentation}
\label{sec:seg}
After collecting $N$ observations of an object from diverse viewpoints, our goal is to segment the object from its background. We first aggregate a colorized pointcloud of the region, by unprojecting each frame using its depth observation and pose, then segment the object from its background in 3D. Figure~\ref{fig:model} shows an overview of this process.

\textbf{2D-to-3D unprojection}\quad For the $i$-th view, a 2D pixel coordinate $(u,v)$ with depth $z$ is unprojected and transformed to its coordinate $(X,Y,Z)^T$ in the reference frame:
\begin{equation}
    (X,Y,Z,1) = \mathbf{G}_{i}^{-1} \left(z \frac{u-c_{x}}{f_{x}}, z \frac{v-c_{y}}{f_{y}}, z, 1\right)^{T}
\end{equation}
where $(f_x, f_y)$ and $(c_x, c_y)$ are the focal lengths and center of the pinhole camera model and $\mathbf{G}_i \in SE(3)$ is the camera pose for view $i$ relative to the reference view. This module first unprojects each RGB image $I_i \in \mathbb{R}^{H\times W \times3}$ into a colored pointcloud in the reference frame $P_i \in \mathbb{R}^{M_i \times 6}$ with $M_i$ being the number of pixels with an associated depth value. We concatenate the spatial location with the RGB color, forming an aggregated colorized scene point cloud. The aggregated pointcloud $P$ can be partitioned into three sets: the foreground set $P_{fg}$, the background set $P_{bg}$, and the unknown set $P_{unk}$:
\begin{equation}
    P = \bigcup\limits_{i=0}^{N-1} P_{i} = P_{fg} \cup P_{bg} \cup P_{unk},
\end{equation}
The per-view foreground/background masks of the detector provide $P_{fg}$ and $P_{bg}$; the rest of the points are denoted by $P_{unk}$. We find it is helpful to erode the predicted segmentation mask to form the foreground mask, to account for mis-classified points near the boundary. Similarly, we dilate the inverse of the object mask to form the 2D background mask.

\textbf{3D Segmentation}\quad
To label points in $P_{unk}$, we apply a simple yet effective two-stage segmentation method. In the first stage, based on the information available in $P_{fg}$ and $P_{bg}$, we train a unary model to classify whether a point belongs to the foreground or background. We then ask the unary classifier to predict the log probability of \textit{all} points, including the unlabelled ones in $P_{unk}$. In the second stage, we apply a fully connected conditional random field (CRF)~\cite{Krahenbuhl2011CRF} to refine the segmentation. 
Each node in the CRF model is a point in $P$ and its unary potential is initialized with the log probabilities from the unary classifier. To inject spatial information into the CRF model, we add a pairwise potential between every point pair $(P^{(a)}, P^{(b)})$ in $P$:\\
\begin{equation}
    \psi_{p}\left(P^{(a)}, P^{(b)}\right)=\mu\left(X^{(a)}, X^{(b)}\right) k(P^{(a)}, P^{(b)})
\end{equation}
where $X^{(a)}$ and $X^{(b)}$ are the first three dimensions of $P^{(a)}$ and $P^{(b)}$ that corresponds to their spatial locations, $\mu$ is the label compatibility function from the Potts Model, and $k$ is a contrast-sensitive potential given by the combination of an appearance kernel and a smoothness kernel (both parameterized by a Gaussian kernel). In our experiments, we use a support vector machine (SVM) as the unary classifier for its efficiency. We additionally experimented with a ``deep'' method for this task, but found this simple two-stage ``shallow'' method to be superior in both accuracy and runtime (see supp.). The output of this process is a high-quality 3D pointcloud segmentation $P_{seg}$, distinguishing the object from its local background. 

\subsection{Supervising with Self-Generated Labels}

After generating 3D pointcloud segmentations of the found objects, our goal is to distill this knowledge into the weights of a neural network. To do this, we treat the estimated 3D segmentations as pseudo-labels, and supervise standard deep architectures to mimic the labels. 

\textbf{3D Detection Training}\quad We compute 3D boxes to encapsulate the 3D segmentations, to match the training format of modern 3D object detectors. Specifically, we compute the height of the box with a min and max on the vertical dimension of the foreground, then project down to a floor plane, and compute a minimum area rectangle to obtain the width, depth and rotation about the vertical axis. We then train a standard object detector on this data from random initialization. Our experiments show that this simple self-supervision scheme outperforms a state-of-the-art self-supervised 3D detection method by a large margin and achieves performance comparable to a supervised detector with the same architecture.


\textbf{2D Detection Training}\quad Since our estimated segmentations are in 3D, and since we have the source views and poses, we are able to create 2D pseudo-labels as well. 
We produce 2D pseudo-labels by re-projecting $P_{seg}$ to all views. For a point $(X,Y,Z)^T$ in $P_{seg}$, we can get its 2D pixel coordinate in the $i$-th frame with:
\begin{equation}
    (u,v)^T = \mathbf{G}_{i} \left(f_{x} \frac{X}{Z}+c_{x}, f_{y} \frac{Y}{Z}+c_{y}\right)
\end{equation}
The reprojected points in $P_{seg}$ are sparse in 2D, so we fit a concave hull to convert them into a connected binary mask.  
Our experiments show that these pseudo-labels are lower quality than ground truth labels, but still provide a valuable boost to a pre-trained detector. 
\section{Experiments}
\label{sec:exp}
\subsection{Datasets}

\textbf{Environments}\quad 
We test our method in an indoor and outdoor environment. We use the CARLA simulator~\cite{Dosovitskiy17CARLA} as the outdoor environment, which renders realistic urban driving scenes. We use the Habitat simulator~\cite{habitat19iccv} with the Replica dataset~\cite{replica19arxiv} as the indoor environment, which contains high quality and realistic reconstructions of indoor spaces.
The Replica dataset consists of 18 distinct indoor scenes, such as offices, hotels, and apartments. We split the scenes into disjoint sets such that there are 10 for training, 4 for validation, and 4 for testing. In our self-supervised data collection, we capture 25 views in each episode, resulting in 17k images for training, 1k for validation, and 2.3k for testing. The CARLA driving scenes consist of five distinct towns. We again split them into 3 towns for training, 1 for validation, and 1 for testing. In our self-supervised data collection, we capture 25 views in each episode. We have 5.3k images for training, 1.8k for validation, and 1.8k for testing. 

\textbf{Objects}\quad For CARLA, we randomly spawn two vehicles in the scene for each episode. Since CARLA has the same semantic label for all vehicles, we consider detection of all vehicle classes in the COCO dataset~\cite{coco2014eccv} during evaluation. For Replica, we keep the default layout of objects in each scene. We consider a subset of object categories based on the following standards: (1) the category is shared between COCO and Replica, and 2) enough instances (more than 10) occur in the dataset. This includes chair, couch, plant, tv and bed. Although our method does not necessitate choosing categories shared between COCO and Replica, it is essential to ground our experimental results in comparison to fully supervised methods that require ground truth annotations for training. We also show results on novel categories not included in COCO.

\subsection{Implementation Details}

\textbf{2D Object Detection}\quad 
For the pre-trained 2D detector, we use a Mask-RCNN~\cite{He2017MaskRCNN} with FPN~\cite{Lin2017FPN} using ResNet-50 as the backbone, pre-trained on the COCO dataset. We fine-tune it on the 2D pseudo-labels from the training set for 100k iterations. To compare, we also fine-tune the detector on the same images but with ground truth labels. In both settings, we use a learning rate of $0.001$ and a batch size of $12$. For selecting the best model, we compute its mAP on a validation set at IoU threshold of 50\% every 5000 iterations. We use the Mask-RCNN implementation from Detectron2~\cite{wu2019detectron2}, keeping all other hyperparameters as default.

\begin{figure}[t]
\begin{center}
\includegraphics[width=\linewidth]{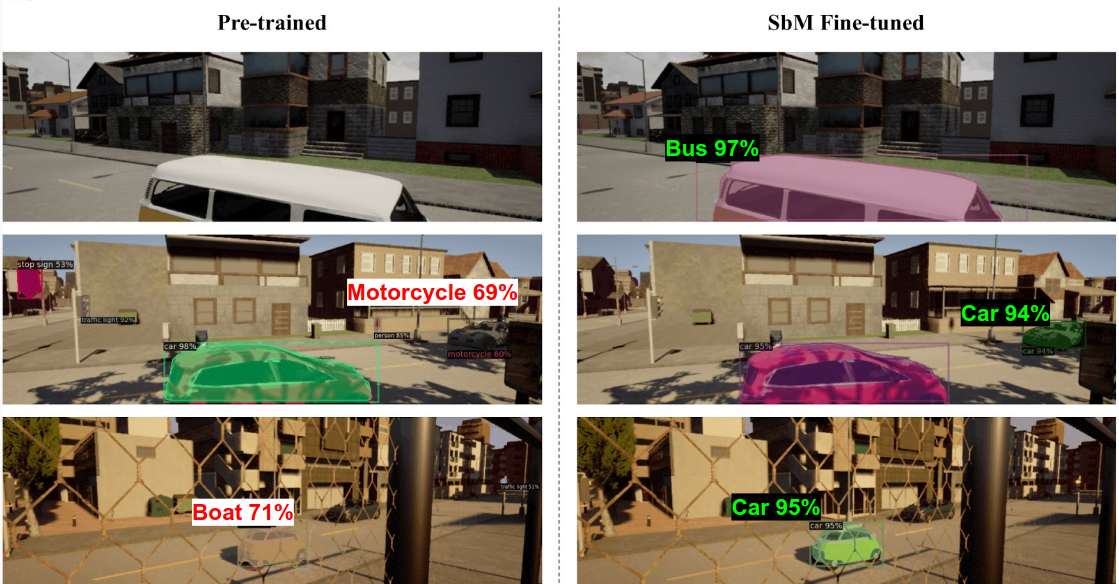}
\end{center}
   \caption{\textbf{Visualizations of 2D detector performance on the CARLA test set.} We show paired qualitative examples of the detections of pre-trained 2D detector (left) and SbM fine-tuned 2D detector (right) on the CARLA test set. The improvements are shown in larger fonts for better visibility. The pre-trained detector misses objects and classifies the object as the wrong category, while the fine-tuned model produces accurate class predictions, bounding boxes, and semantic masks.}
\label{fig:carla_qual}
\end{figure}

\begin{table}[t]
\centering
\begin{tabular}{@{}cccc@{}}
\toprule
mAP@IoU & Method                     & Train & Test  \\ \midrule
\multirow{3}{*}{0.5} & Pre-trained & 68.05  & 68.23   \\
& SbM Labels &  86.88 & 81.81  \\
& SbM fine-tuned & - & 80.15 \\
\cline{2-4}
& GT fine-tuned    &  - & 93.76  \\ 
\hline
\multirow{3}{*}{0.3} & Pre-trained & 73.09 & 75.55  \\
 & SbM Labels  & 92.93 & 92.49 \\
 & SbM fine-tuned & - & 88.84 \\
 \cline{2-4}
& GT fine-tuned  & -  & 94.71  \\ \bottomrule \\
\end{tabular}
\caption{\textbf{2D object detection performance comparison on CARLA test set} Fine-tuning 2D detector on self-supervised SbM labels increases pre-trained models performance taking its performance closer to supervised fine-tuning.}
\label{tab:carla}
\end{table}

\begin{figure*}[t]
\begin{center}
\includegraphics[width=0.98\linewidth]{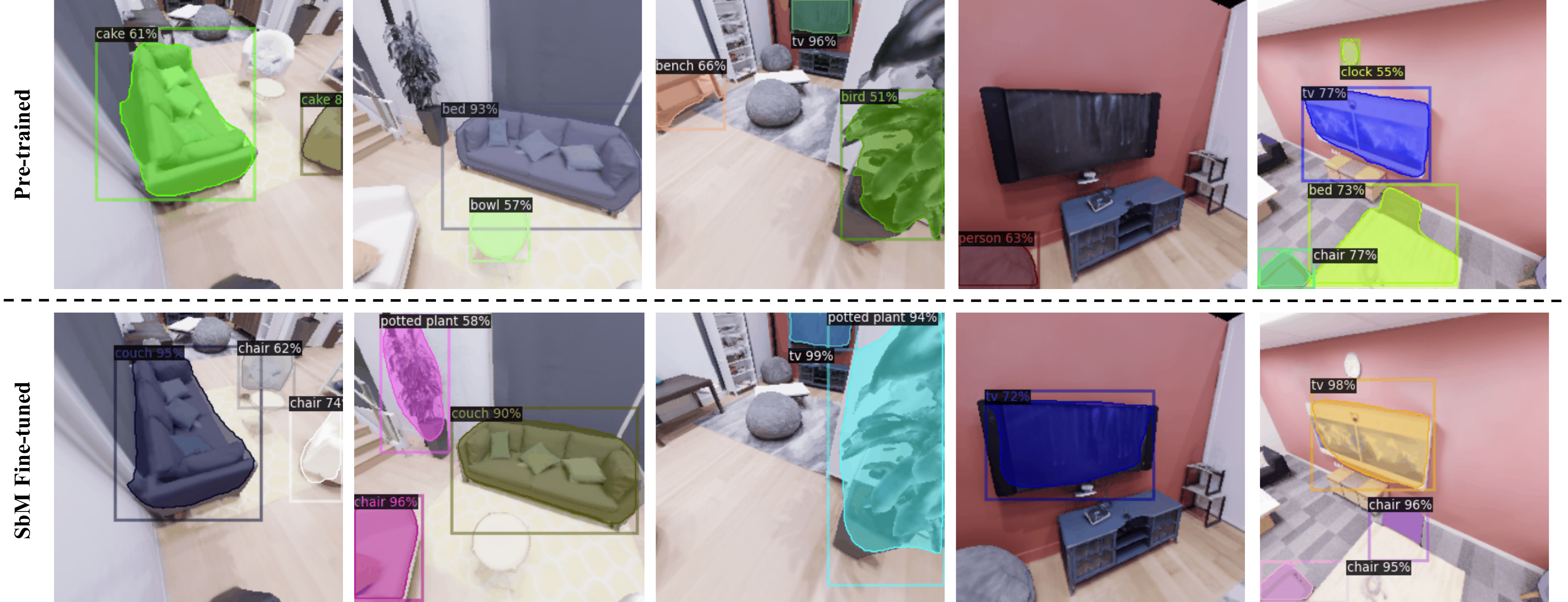}
\end{center}
   \caption{\textbf{Visualizations of 2D detector performance on the Replica test set.} We show paired qualitative examples of the predictions of the pre-trained 2D detector (top) and the SbM fine-tuned 2D detector (ours) (bottom). The pre-trained detector misses objects and classifies the object as the wrong category, while the fine-tuned model produces accurate class predictions, bounding boxes, and semantic masks.}
\label{fig:replica_qual}
\end{figure*}

\textbf{3D Object Detection}\quad
We use the frustum PointNet model~\cite{qi2017frustum} with PointNet~\cite{qi2016pointnet} backbone on CARLA. The original frustum PointNet model uses ground truth 2D bounding boxes and camera pose to define a 3D frustum search space and then performs 3D segmentation on it using a PointNet-based architecture, and uses ground truth 3D boxes for supervision. In our experiment, we use the 2D and 3D bounding boxes generated from our self-supervised 3D segmentation. To compare, we also train the same network using ground truth 2D and 3D bounding boxes. For both settings, we train it using a learning rate of $0.001$ and a batch size of $32$  until convergence. For selecting the best model, we use the validation loss curve. We test both models on a new unseen town and compare their performance.
\subsection{2D Object Detection}
We analyze our method for improving 2D object detection by asking two questions: (1) do our pseudo-labels outperform the detector on which they are based? (2) does fine-tuning the object detector on the pseudo-labels improve the detector's performance on unseen scenes? Experiments in CARLA and Replica show that the answer to both is ``yes".

\textbf{CARLA}\quad
The performance of our method, the pre-trained detector, and the detector fine-tuned on ground truth data is shown in Table~\ref{tab:carla}. We report mAP at IoU of $0.5$ and $0.3$ using the PascalVOC mAP implementation of Padilla et al.~\cite{map}. At training time, we investigate the setting where the embodied agent is free to move around, obtain observations, and use SbM to generate predictions for all views. We observe that pseudo-labels generated by SbM have much better performance than the pre-trained detector outputs. This shows that moving and performing segmentation in the 3D space helps the detector see better, when compared to treating multi-view images as individual observations. We also finetune the detector on the SbM pseudo-label dataset generated from training set. At test time, the SbM fine-tuned model is deployed in unseen environments where only a single RGB image is given as input. Results show that the detector fine-tuned with SbM outperforms the pre-trained detector by a large margin. This indicates that we can improve the detector's performance in unseen environments with no additional human labels. 
Figure~\ref{fig:carla_qual} shows qualitative comparisons of the detections of the pre-trained detector and the detector fine-tuned by SbM pseudo-labels.

\begin{table*}[hbt!]
\begin{center}
\begin{tabular}{@{}cccccccc|c@{}}
\toprule
mAP@IoU & Method & Bed   & Chair & Couch & Table & Plant & TV    & Avg \\ \midrule
\multirow{3}{*}{0.5} & Pre-trained & \textbf{7.50}  & 11.08 & 17.20 & \textbf{7.09}  & 20.44        & 46.79 & 18.35 \\
& SbM (ours) & 6.08  & \textbf{21.41} & \textbf{39.67} & 4.12         & \textbf{27.15}        & \textbf{58.78} & \textbf{26.20} \\
\cline{2-9}
& SbM-ws (ours) & 7.34  & 40.53 & 58.33 & 38.33 & 64.68  & 58.23 & 44.57  \\
\hline
\multirow{2}{*}{0.3} & Pre-trained & 8.12 & 13.18 & 17.97 & 7.41 & 48.28    & 46.79 & 23.62\\
 & SbM (ours)           & \textbf{10.04} & \textbf{31.76} & \textbf{45.63} & \textbf{8.30}         & \textbf{66.99}        & \textbf{66.04} & \textbf{38.12}\\ 
 \cline{2-9}
& SbM-ws (ours) & 39.03  & 58.93 & 82.37 & 59.74 & 82.85 & 75.87   & 66.47 \\
\bottomrule \\
\end{tabular}
\caption{\textbf{2D object detection performance of the pre-trained detector vs self-supervised SbM vs weakly supervised SbM on the Replica \textit{training set}.} Self-Supervised SbM consistently outperforms the pre-trained detector across most categories. Providing weak supervision (a single-view ground truth annotation) to SbM increases its performance significantly on all categories.}
\label{tab:replica_train}
\end{center}
\end{table*}

\begin{table*}[hbt!]
\begin{center}
\begin{tabular}{@{}cccccccc|c@{}}
\toprule
mAP@IoU & Method                     & Bed   & Chair & Couch & Table & Plant & TV    & Avg \\ \midrule
\multirow{3}{*}{0.5} & Pre-trained & \textbf{15.18}  & 21.51 & \textbf{23.54} & 2.37         & 11.74        & 43.71  & 19.67 \\
& SbM Fine-tuned (ours)            & 5.57  & \textbf{36.19} & 18.86 & \textbf{8.50}         & \textbf{37.34} & \textbf{57.85} & \textbf{27.38} \\
\cline{2-9}
& GT Fine-tuned             & 27.20  & 53.56 & 48.65 & 26.99         & 35.04        &  58.28   & 41.62 \\ 
\hline
\multirow{3}{*}{0.3} & Pre-trained & \textbf{27.71} & 22.95 & \textbf{25.83} & 2.80 & 19.79   & 43.71 & 23.79 \\
 & SbM Fine-tuned (ours)       & 10.55  & \textbf{45.60} & 21.17 & \textbf{8.82} & \textbf{40.80}         & \textbf{57.85}      & \textbf{30.79}  \\
 \cline{2-9}
& GT Fine-tuned             & 38.25  & 60.15 & 52.84 & 28.65 & 42.59        & 58.28  & 46.79 \\ 
\bottomrule \\
\end{tabular}
\caption{\textbf{2D object detection performance of pre-trained, SbM fine-tuned (ours), and ground truth fine-tuned detector on the Replica \textit{test set}.} Training on SbM-generated pseudo-labels improve the detector performance on the test set by a large margin.} 
\label{tab:replica_test}
\end{center}
\end{table*}

\textbf{Replica}\quad 
\begin{table}[!htb]
\centering
\begin{tabular}{ccc}
\toprule
Method & mAP@0.5 & mAP@0.3 \\\midrule
Pre-trained & 21.36 & 26.14 \\
SbM Labels w/ noise (ours) &  \textbf{23.15} & \textbf{33.68} \\
\cline{2-3}
SbM Labels w/o noise (ours) &  26.20 & 38.12 \\
\bottomrule \\
\end{tabular}
\caption{\textbf{Pseudo-label accuracy with pose noise in the Replica training set.} We show that actuation noise weakens the data collection, yet our method is still able to produce pseudo-labels that are better than the pre-trained detectors' predictions. }
\label{tab:noise}
\end{table}
The performance of our pseudo-labels on the training set is shown in Table~\ref{tab:replica_train}. We observe that our pseudo-labels are more accurate than the pre-trained detector on most classes, indicating that moving around helps generate better labels. Note that the performance on ``table" and ``bed" category is very low for both the pre-trained detector and SbM. This is because these classes in COCO are visually very different from the respective class in Replica (see supp.). The performance comparison of the pre-trained, SbM fine-tuned, and ground truth fine-tuned detectors on the test set is shown in Table~\ref{tab:replica_test}. The SbM fine-tuned detector overall outperforms the pre-trained detector by a large margin on most categories. In Figure~\ref{fig:replica_qual}, we also present qualitative comparisons of the detections of the pre-trained detector and the detector fine-tuned by SbM pseudo-labels. This confirms that fine-tuning on pseudo-labels generated by moving around makes the detector robust to viewpoint changes. Surprisingly, the performance of couch decreased after fine-tuning even though the pseudo-label mAP of couch is higher than pre-trained detector on the training set. This is because the pre-trained detector often recognizes couch as bed with high confidence, thus corrupting the pseudo-labels (see supp.). This reveals a limitation of our method, which can be mitigated by using context-aware detectors~\cite{Marino2017themoreyouknow, Wang2018zero} or by providing weak supervision, as in Section~\ref{sec:weak_novel}. In addition, we show in Table~\ref{tab:noise} that this improvement over the baseline is maintained even under realistic actuation noise (see supp.). 

\subsection{3D Object Detection}
\begin{figure}[t]
\begin{center}
\includegraphics[width=\linewidth]{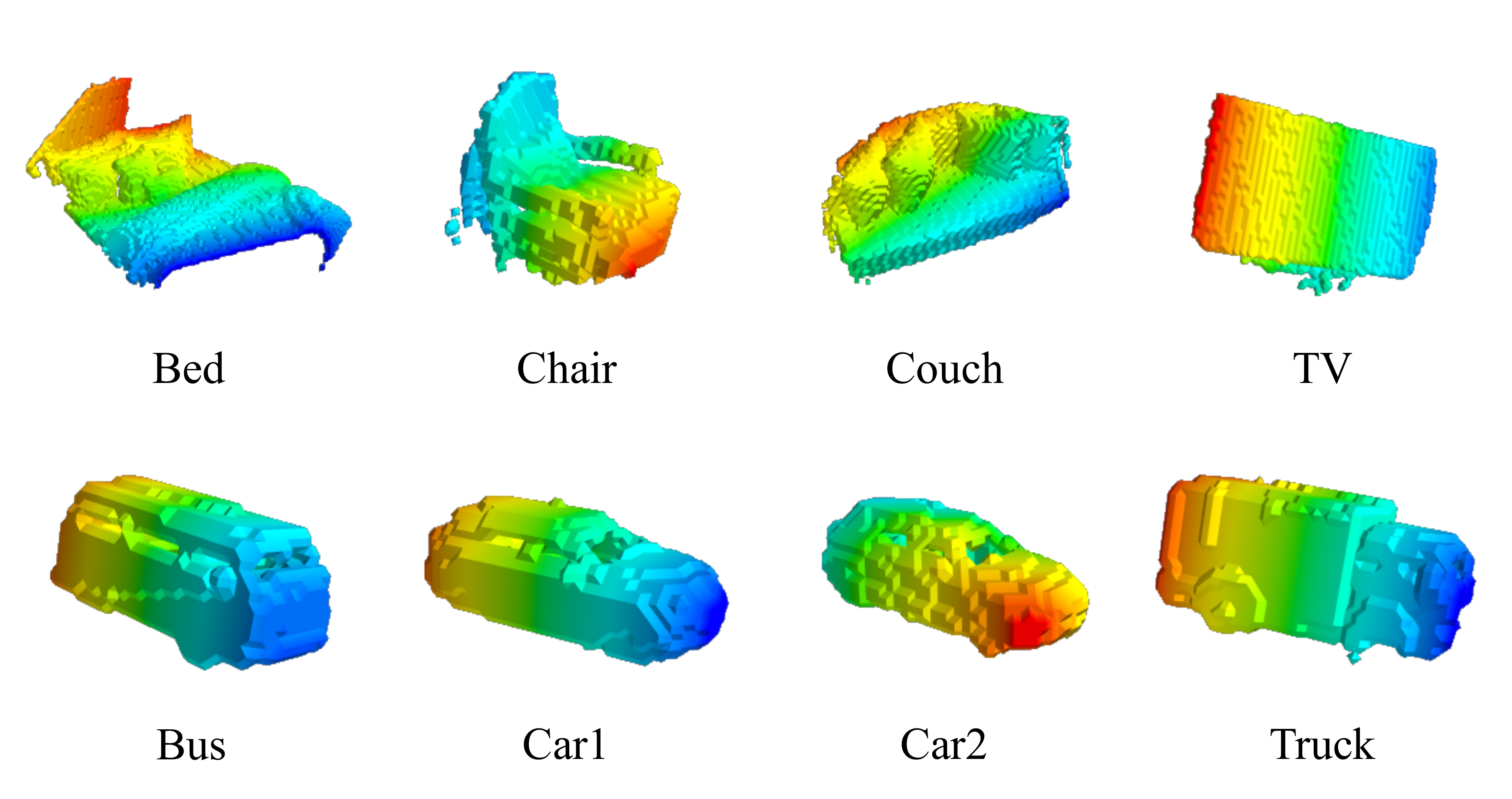}
\end{center}
   \caption{\textbf{Visualizations of 3D object segmentation.} We show colorized visualisations of the voxelized 3D segmentations of our method, on Replica (top) and CARLA (bottom).}
\label{fig:3d_segs}
\end{figure}

Can we train a 3D object detector self-supervised without any ground truth data? To answer this question, we compare the two versions of frustum PointNet: one trained on SbM's self-supervised 3D and 2D labels (Figure~\ref{fig:3d_segs}), and the other trained on ground truth 3D and 2D labels. We also compare our method with the semi-supervised LDLS~\cite{Wang2019LDLS} method. The experiments are conducted in CARLA.

\begin{figure}[!htb]
\begin{center}
\includegraphics[width=\linewidth]{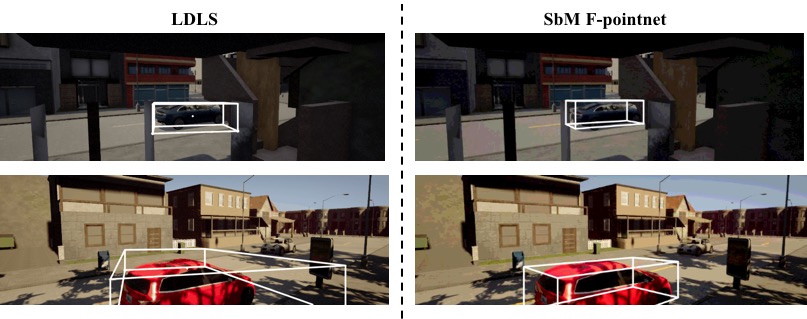}
\end{center}
   \caption{\textbf{Visualizations of 3D object detection.} We show paired examples of the results of LDLS~\cite{Wang2019LDLS} (left) and SbM-trained frustum PointNet (ours, right) on the CARLA test set. While LDLS estimates the bounding box roughly, the SbM-trained frustum PointNet is able to obtain much tighter and better-oriented boxes.}
\label{fig:carla_3d_qual}
\end{figure}

\begin{table*}[t]
\begin{center}
\begin{tabular}{@{}ccccccc@{}}
\toprule 
mAP@IoU              & Method Name   & Cushion   & Nightstand & Shelf & Beanbag & Avg\\ 
\hline
\multirow{2}{*}{0.5} & SbM-ws Trained         & \textbf{93.62} & \textbf{81.25} & \textbf{24.38} & 82.18 & \textbf{70.35} \\
                     & Limited GT Trained         &  87.24  & 79.79 & 16.40 & \textbf{88.77} & 68.04\\
\hline
\multirow{2}{*}{0.3} & SbM-ws Trained        & \textbf{94.23} & \textbf{81.25} & \textbf{25.61} & 82.18 &  \textbf{70.81} \\
                     & Limited GT Trained          & 87.24  & 79.79 & 16.40 & \textbf{88.77} & 68.04\\
                     
\bottomrule \\
\end{tabular}
\caption{\textbf{SbM-ws labels can be used to train detectors on novel categories.} We compare the performance of the detector trained on labels produced by SbM-ws with the detector trained on ground truth labels. The results show that the detector trained on SbM-ws labels outperforms the on trained on limited ground truth labels.}
\label{tab:novel_test}
\end{center}
\end{table*}

\begin{table}[t]
\centering
\begin{tabular}{cc}
\toprule
 Method & mAP@0.25 \\ \midrule
LDLS~\cite{Wang2019LDLS}  & 44.03 \\
SbM Self-Sup. F-PointNet (ours)  & \textbf{83.87} \\
\cline{2-2}
Supervised F-PointNet  &  85.06   \\ \bottomrule \\
\end{tabular}
\caption{\textbf{Fine-tuning with SbM labels outperforms self-supervised LDLS.} 3D object detection performance of LDLS~\cite{Wang2019LDLS}, frustum PointNet trained on SbM segmentations, and GT-trained frustum PointNet on the CARLA test set.}
\label{tab:carla_3d}
\end{table}

Table ~\ref{tab:carla_3d} shows the test set performance of LDLS~\cite{Wang2019LDLS}, frustum PointNet trained on SbM segmentations, and frustum PointNet trained on ground truth. Our self-supervised frustum PointNet model outperforms LDLS significantly. Our model also achieves close performance to the fully supervised model. We show qualitative examples of the 3D detections from LDLS and SbM fine-tuned frustum PointNet in Figure~\ref{fig:carla_3d_qual}. This demonstrates that the 3D segmentation labels produced by SbM are high quality and could be successfully used to train state of the art 3D detection models without ground truth 3D annotations.


\subsection{Weakly-Supervised Novel Object Detection}
\label{sec:weak_novel}
In this section, we ask two questions. (1) As previously noted, moving around does not help much in cases where the pre-trained detector performs poorly on all views. Can we generate higher-quality labels if provided weak supervision? (2) Since embodied agents typically encounter novel objects while exploring, it would be useful if those objects could be learned with a small number of human annotations. Can we learn those categories with weak supervision?

We show that our method can perform pseudo-label generation when we have weak ground-truth 2D annotations, enabling us to generate high quality labels for non-COCO instances, as well as for categories where the pre-trained detector fails. In our experiments, we only provide a ground truth annotation on 1 view for each object out of the 25 available views, making the label propagation procedure weakly supervised in 2D. We denote this setup as SbM-ws.

To answer the first question, we report the label quality on the training set in Table~\ref{tab:replica_train}. We observe that with a single labelled view, the pseudo-label quality is better than both the pre-trained detector and self-supervised SbM by a large margin. This suggests that our method can generate much better pseudo-labels with an improved pre-trained detector.

To answer the second question, we provide weak supervision for four novel non-COCO objects: Cushion, Nightstand, Shelf and Beanbag. To make a fair comparison, we compare the performance of the detector fine-tuned on pseudo-labels with the detector fine-tuned on the same provided ground truth labels. The results of the experiment are shown in Table~\ref{tab:novel_test}. We see that the detector trained on SbM pseudo-labels outperforms the detector trained on the limited ground truth data. This is because our method effectively creates more training data by propagating the weak supervision to more views.

\section{Conclusion}
While visual recognition systems trained on large internet data have shown great advancements, they still require a lot of additional human annotations to work well on novel domains, unusual poses, or heavy occlusion conditions. Motivated by how humans learn, we push for an active agent that can move in the environment, discover objects, and generate its own pseudo-labels for self-supervision. 
In both indoor and outdoor settings, we show that our method significantly improves the performance of a pre-trained detector in test environments. Moreover, we show that our method can be used to train a 3D detector without any human-provided 3D annotations. Our experiments with simple exploration policies and realistic actuation noise show promising results for real-world conditions. We believe that active visual learning remains an important problem for future work. Our method assumes that the pre-trained detector makes correct high-confidence predictions for at least some of the available views. When this assumption does not hold (and the model makes high-confidence \textit{incorrect} predictions), the pseudo-labels are likely to be corrupted with bad information, leading to ineffective self-supervision.
It may be helpful to use contextual or common-sense cues to improve recognition or to find unusual objects \cite{torralba2003contextual}. 
Finally, while we try to imitate real-world situations through our problem design and realistic simulators, applying this method on real-world data is a direct avenue for future research.    


{\small
\bibliographystyle{ieee_fullname}
\bibliography{egbib}
}
\clearpage
\appendix
\section{Overview}
The structure of this supplementary file is as follows: Section~\ref{sec:results} provides a more detailed analysis of the method by performing ablation and comparison experiments. Section~\ref{sec:add_vis} includes additional visualizations to help better understand the both method's strengths and limitations. We also include a video named ``\texttt{sup\_vid\_3091.mp4}'' with additional 3D visualizations and urge the reviewers to refer that as well.

\section{Experiments and Ablations}
\label{sec:results}
\subsection{Pre-trained Detector Quality}
\begin{figure*}[!htb]
\begin{center}
\includegraphics[width=\linewidth]{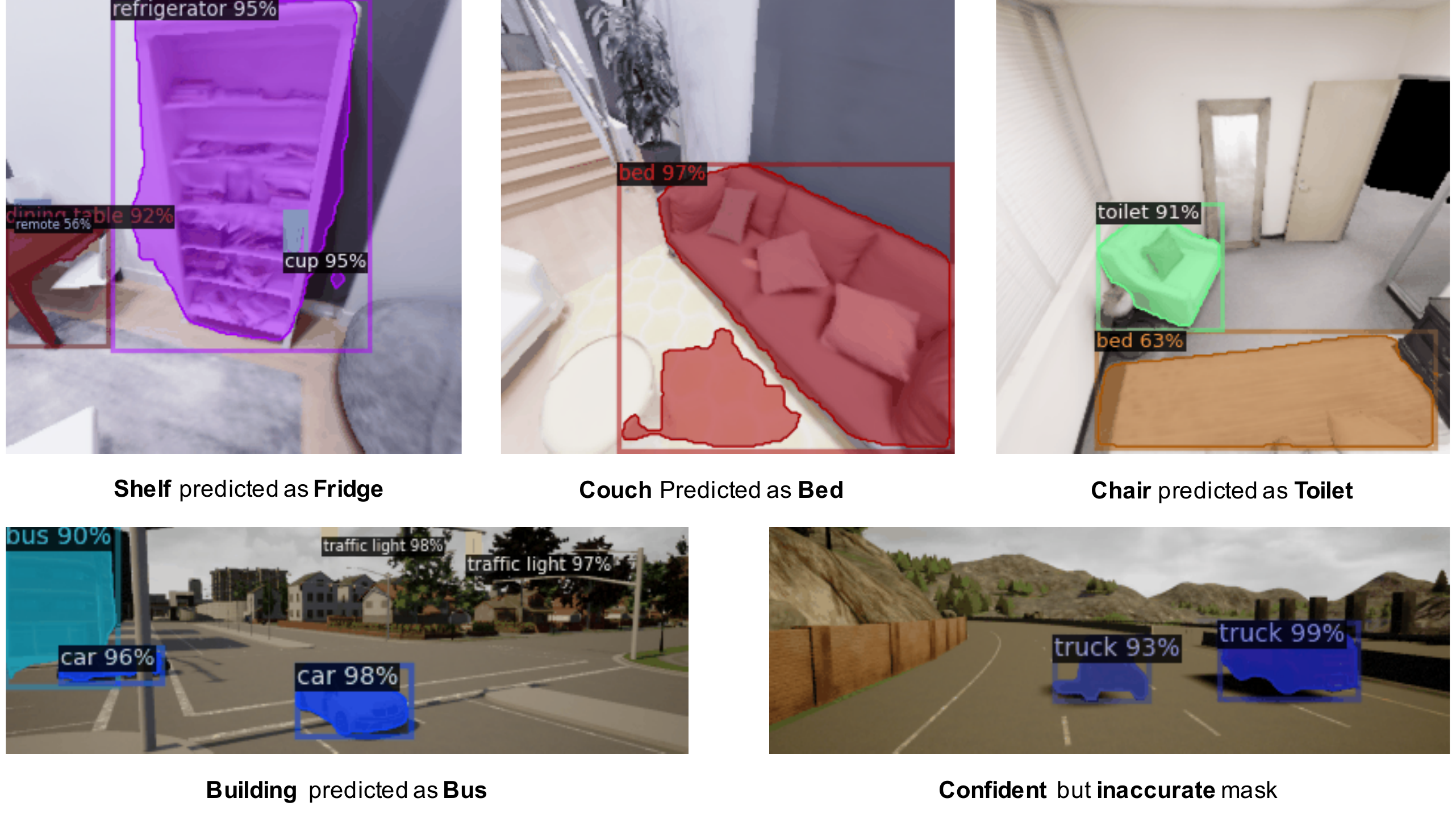}
\end{center}
   \caption{\textbf{Incorrect detections by pre-trained detector with high confidence.} We show three examples where the pre-trained detector incorrectly classify the object with high confidence.}
\label{fig:pretrainedwrong_pred}
\end{figure*}
As mentioned in the paper, a possible limitation of our method is that in order to propagate high-quality labels, the pre-trained detector must detect objects correctly with high confidence. In novel environments and viewpoints, the pre-trained detector sometimes detects wrong objects with high confidence, as shown in Figure~\ref{fig:pretrainedwrong_pred}. In this experiment, we ask the following question: is the confidence score of the pre-trained detector reliable to some degree?

To investigate this question, we set different confidence score thresholds $\theta$ for the pre-trained detector and report their precision, recall, and mAP@IoU=0.5 on CARLA training set. The results are shown in Table~\ref{tab:precision_recall}. We observe that although the detector's confidence does not serve as a perfect cue all the time, it is calibrated enough for our method's assumption to hold. Therefore, by setting a high confidence threshold we are able to obtain high precision detections. Our 3D segmentation then propagates the high precision detections to all views, resulting in pseudo-labels with both high precision and high recall, as shown in the last column in Table~\ref{tab:precision_recall}. We also note that recent works~\cite{EBMclassifier_iclr_2020} have shown promising results in training classifiers that are well-calibrated while preserving the performance.
\begin{table}[!htb]
\centering
\begin{tabular}{cccc}
\toprule
Setting & Precision & Recall & mAP@0.5 \\\midrule
Pre-trained, $\theta=0.5$ & 83.97 & 81.91 & 68.05 \\
Pre-trained, $\theta=0.6$ & 85.66 & 80.43 & 68.96 \\
Pre-trained, $\theta=0.7$ &  87.46 & 76.74 & 67.60 \\
Pre-trained, $\theta=0.8$ &  89.58 & 70.71 & 64.06 \\
Pre-trained, $\theta=0.9$ &  92.41 & 60.99 & 57.23 \\
Pseudo-labels, $\theta=0.9$ & 92.92 & 92.76 & 86.88 \\
\bottomrule \\
\end{tabular}
\caption{\textbf{The pre-trained detector is overall well-calibrated.} We set different confidence score thresholds for the pre-trained detector and report their precision, recall, and mAP@IoU=0.5 on CARLA training set.}
\label{tab:precision_recall}
\end{table}

\subsection{Design Choice of 3D Segmentation}
In our method, we employ a simple yet effective two-stage 3D segmentation method for its fast runtime and strong performance. In this experiment, we ask the following question: will an alternative ``deep'' segmentation method improve the performance, without sacrificing runtime? 

To investigate this question, we replace the unary classifier in our segmentation module with a PointNet~\cite{qi2016pointnet}. For each scene, an input sample to the PointNet is a random subset of $P \in \mathbb{R}^{M_i \times 6}$ with $M_i$ being the number of pixels with an associated depth value. We sample randomly while making sure that in each sample there exist both points from $P_{fg}$ and $P_{bg}$. During training, loss is only computed for points from $P_{fg}$ and $P_{bg}$. We also ablate the effectiveness of our refinement stage with CRF, reporting the performance both with and without CRF refinement. We perform this experiment on CARLA training set to evaluate the resulting pseudo-label quality. The results are shown in Table~\ref{tab:pointnet_comparison}. We observe that PointNet improves marginally compared to the SVM alone, while being about 280$\times$ slower. Adding CRF refinement stage significantly improves SVM, while only marginally improving PointNet. Therefore, we show that our two-stage 3D segmentation method is simple, fast, and effective. It is important to note that training a single PointNet on many $P$'s from many episodes might yield better performance for such a deep model, but that requires many more episodes of data collection. In contrast, our method operates on each episode separately and generates high-quality pseudo-labels even in the low-data regime.
\begin{table}[!htb]
\centering
\begin{tabular}{cccc}
\toprule
Method & mAP@0.5 & mAP@0.3 & runtime (s)\\\midrule
SVM & 67.57 & 87.44 & 1.18\\
SVM+CRF & 86.88 & 92.93 & 1.46 \\
PointNet & 69.52 & 84.47 & 332.04 \\
PointNet+CRF & 71.85 & 85.22 &  332.32 \\
\bottomrule \\
\end{tabular}
\caption{\textbf{Performance and runtime comparison with PointNet.} We perform comparison between our two-stage segmentation method and PointNet, as well as ablate the effectiveness of CRF as a refinement module.}
\label{tab:pointnet_comparison}
\end{table}

\subsection{Realistic Noise on CARLA}
\label{sec:carla_noise}
In Section 4.3 of the main paper, we show that our method can work even when realistic actuation noise is added to the Replica dataset. In this experiment, we add similar realistic actuation noise in CARLA and repeat our experiments. The pseudo-label quality is evaluated on the CARLA training set, shown in Table~\ref{tab:carla_noise}. We observe that our method is still able to produce significantly better pseudo-labels compared to the pre-trained detector's predictions, even under realistic noise settings. In addition, we show qualitative segmentation results in the included video.

\begin{table}[!htb]
\centering
\begin{tabular}{ccc}
\toprule
Method & mAP@0.5 & mAP@0.3 \\\midrule
Pre-trained & 68.05 & 73.09 \\
SbM Labels w/ noise (ours) &  \textbf{75.75} & \textbf{91.44} \\
\cline{2-3}
SbM Labels w/o noise (ours) &  86.88 & 92.93 \\
\bottomrule \\
\end{tabular}
\caption{\textbf{Pseudo-label accuracy with pose noise on the CARLA training set.} We show that actuation noise weakens the data collection, yet our method is still able to produce pseudo-labels that are better than the pre-trained detector's predictions.}
\label{tab:carla_noise}
\end{table}

\subsection{Effect of the Number of Views}
In our experiments in the main paper, we set the number of views to $N=25$ in all cases. In this experiment, we ablate the effect of adding more views to the scene point cloud on the performance of pseudo-labels. The quality of pseudo-labels is evaluated on the CARLA training set. The results are shown in Table~\ref{tab:vary_views}. We observe a steady increase in accuracy initially, but it starts to saturate with an increased number of images. Since our method require at least 1 confident detection out of all views, scaling the number of views initially increases the probability of finding a confident detection. After a certain number of views, confident views are captured more often than not, resulting in a diminishing return in accuracy of pseudo labels. However, we expect the performance of MaskRCNN fine-tuned with our pseudo labels to increase monotonously with increasing views simply because of increased training dataset size. 

\begin{table}[htb!]
\centering
\begin{tabular}{cccccccc}
\toprule
\# views & 2 & 5 & 10 & 15 & 20 & 25\\ \hline
mAP  & 64.9 & 75.9 & 84.4 & 85.5 & 87.2 & 86.9 \\
\bottomrule \\
\end{tabular}
\caption{\textbf{Effect of the number of views} We report mAP@IoU=0.5 on the CARLA training set, when we use 2, 5, 10, 15, 20, 25 views in each episode. When the number of views are less than 25, we randomly sample without replacement.}
\label{tab:vary_views}
\end{table}

\subsection{Multi-view Aggregation for Segmentation}
Since in an episode we have multiple views of the objects, it is often likely that multiple confident detections of the same object from different views exist. Therefore, another natural question that may arise in our setup is: does leveraging multi-view information help segmentation? 

In this experiment, we explore a variant of our method that aggregates confident detections from multiple views to obtain the 3D segmentation. However, correspondence does not come for free: we don't know how different detections from different views correspond to one another. To sidestep this limitation and combine multiple detections of the same object from different views, we use a voting mechanism. For the $i$-th object category, we keep a voxel grid $V_i$ with dimensions $X, Y, Z$ of size $X \times Y \times Z$ by voxelizing pointclouds in the reference frame. The confident detections of all instances from the $i$-th category are unprojected, transformed to the reference frame, voxelized, and added into $V_i$. To segment a particular instance for class $i$, we initialize the sets $P_{bg}$ the same way as before, but initialize $P_{fg}$ from the largest connected component in $V_i$ that overlaps with the unprojection of that instance detection. Then, the same two-stage segmentation method is used.

We compare the pseudo-label quality of the method in our main paper with this multi-view aggregation variant on the CARLA training set, and report the results in Table~\ref{tab:multiview_agg}. The results suggest that pseudo-labels from multi-view detection aggregation significantly outperform the detections of the pre-trained detector, but underperforms our original method. Furthermore, the performance drop for mAP@0.5 is larger than the drop for mAP@0.3. We conjecture that this is due to two reasons: (1) to effectively cast votes from multi-view detections, we voxelize the points after they have been transformed to the reference frame, losing detailed information about the point locations; (2) our two-stage segmentation method can already perform robust segmentation of the full 3D object from a small subset of $P_{fg}$ and $P_{bg}$ (from a single detection). 

\begin{table}[!htb]
\centering
\begin{tabular}{ccc}
\toprule
Method & mAP@0.5 & mAP@0.3 \\\midrule
Pre-trained & 68.05 & 73.09 \\
SbM Labels w/ view agg. &  80.29 & 89.26 \\
SbM Labels original &  \textbf{86.88} & \textbf{92.93} \\
\bottomrule \\
\end{tabular}
\caption{\textbf{Pseudo-label accuracy with multi-view aggregation} We show that multi-view detection aggregation significantly improves over the baseline pre-trained detector, but underperforms our original method.}
\label{tab:multiview_agg}
\end{table}

\section{Additional Visualizations}
\label{sec:add_vis}
\subsection{Visually Unmatched Categories}
\begin{figure}[!htb]
\begin{center}
\includegraphics[width=\linewidth]{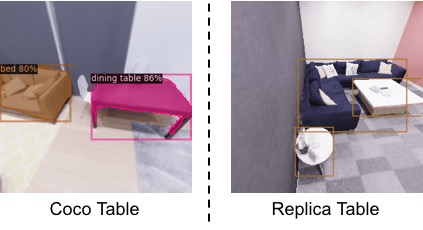}
\end{center}
   \caption{\textbf{Semantically and visually different tables in COCO and Replica.} We show a table (predicted as ``dining table" by the pre-trained MaskRCNN) on the left, and two actual tables in Replica dataset on the right.}
\label{fig:table_comparison}
\end{figure}

As we described in Section 4.3 of the paper, categories that overlap between COCO and Replica are sometimes visually (and even semantically) very different, making it hard to obtain high confidence detections of certain objects. In Figure~\ref{fig:table_comparison}, we show that the tables in COCO and Replica differ significantly in semantic meaning and appearance. In the left image (from Replica), a detector pre-trained on COCO identifies a ``dining table'' with a confidence score of 86\%, while it is not labelled as a table in the Replica annotations. In the right image, the bounding boxes show two tables and a couch annotated in Replica. We observe that these annotated tables are visually very different from the ``dining table" class in COCO.

\begin{figure*}[!htb]
\begin{center}
\includegraphics[width=\linewidth]{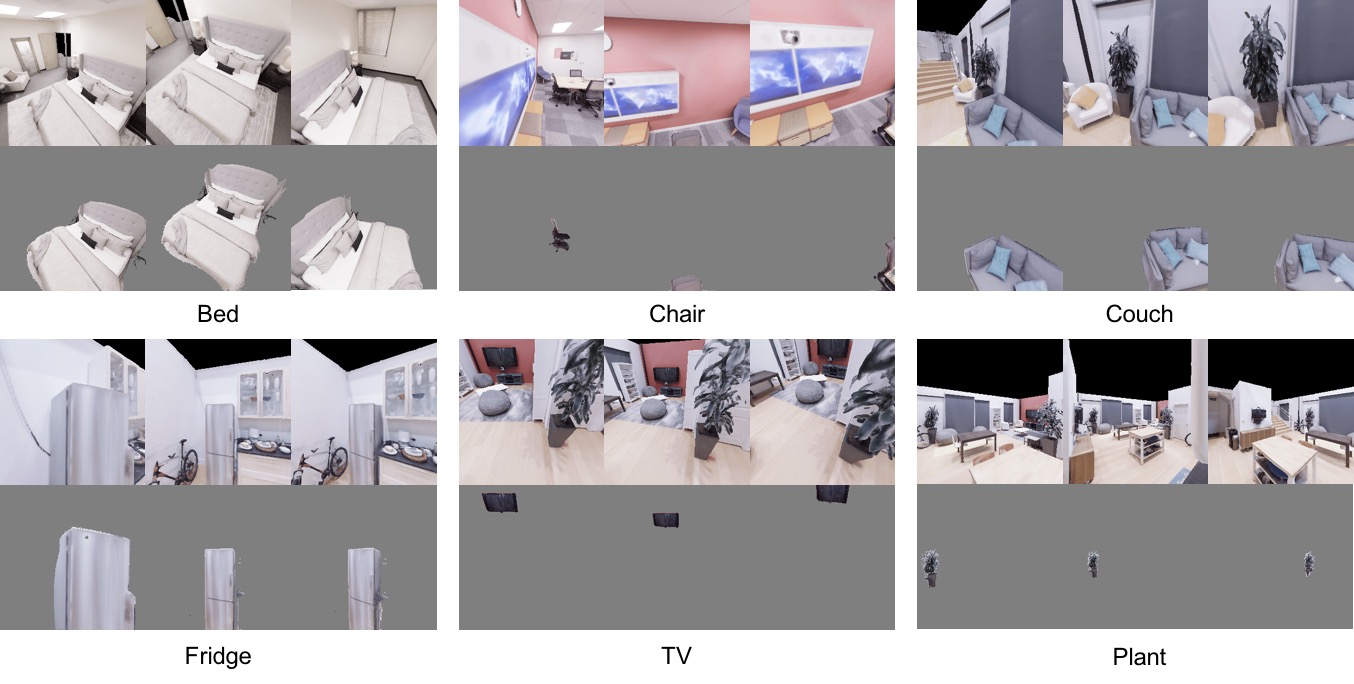}
\end{center}
   \caption{\textbf{Example 2D pseudo-labels reprojected from 3D segmentation.} We show examples of the reprojections of 3D segmentation (which are used as 2D pseudo-labels) on a variety of objects in the Replica dataset.}
\label{fig:reproj}
\end{figure*}
\subsection{Pseudo-label Visualizations} We show visualizations of the 2D pseudo-label masks re-projected from the 3D segmentation for a variety of classes in the Replica dataset in Figure~\ref{fig:reproj}. We can see that the segmentation is complete with sharp borders between foreground and background. In addition, we show qualitative segmentation results in the included video.

\subsection{Weakly-supervised Novel Object Detection}
\begin{figure*}[!htb]
\begin{center}
\includegraphics[width=\linewidth]{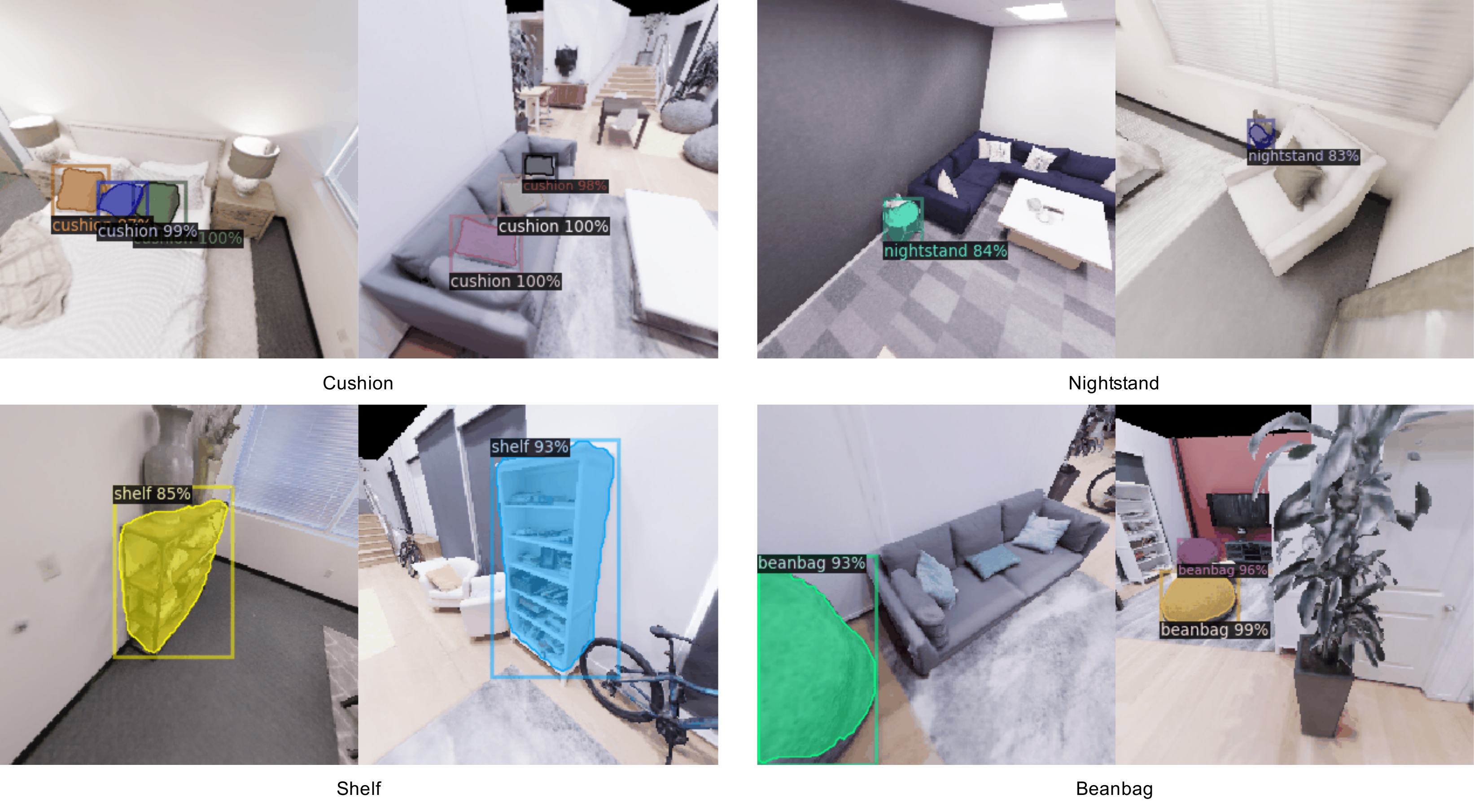}
\end{center}
   \caption{\textbf{Visualizations of the detection results of the detector trained with SbM-ws pseudo-labels} From visualizations of detection results on the test set, we can see that the detector supervised by SbM-ws pseudo-labels generates robust predictions.}
\label{fig:novel_obj_det}
\end{figure*}

In Figure~\ref{fig:novel_obj_det}, we show visualizations of the of 2D detector performance fine-tuned on SbM-ws pseudo-labels. The pseudo-labels are generated with weak supervision (ground truth on 1 view per episode) on novel categories (corresponding to Section 4.5 of the main paper). The detector detects the objects under diverse viewpoints.

\section{Implementation Details}

\subsection{Data Collection Details}
For CARLA, we use a radius of 3.0-15.0 meters from the target object 3D centroid for sampling goal locations for navigation. For Replica, we use a radius of 0.5-3.0 meters. We estimate the 3D centroid of the target detection by unprojecting the median depth value of the masked depth image. We obtain 25 views in each episode (sample 25 goal locations). 

\subsection{Egomotion Error During Realistic Noise}
Section 4.3 of the main paper and Section~\ref{sec:carla_noise} of this supplementary material demonstrate the application of our method with realistic egomotion noise~\cite{chaplot2020neuralslam}. Details of the egomotion noise is described in Section 3.1 of the main paper. Due to the variability in the egomotion estimates obtained from this setting, we apply an additional constraint to remove pairs of frames where the optical flow obtained from the egomotion estimate is not cycle-consistent. This is a well-known “check” for optical flow – if flow does not align when computed in the forward ${t\rightarrow t+1}$ and the backward ${t+1\rightarrow t}$ direction, it is not likely to be correct. We use this strategy to identify exceptionally poor egomotion estimates between frames to remove them. 

We first generate forward flow $\mathbf{w}_{t\rightarrow t+1}$ and backward flow $\mathbf{w}_{t+1\rightarrow t}$ between each pair of views by warping the 3D point cloud from the first view to the second view using the noisy egomotion estimate, and take the delta between 2D points to be the flow. We estimate the rigid motion twice: once using the forward flow, and once using the backward flow (which delivers an estimate of the inverse transform, or backward egomotion). We then measure the inconsistency of these results, by applying the forward and backward motion to the same pointcloud, and measuring the displacement: 
\begin{equation}
XYZ_0^{'} = RT_{10}^{bw}RT_{01}^{fw}(XYZ_0)
\end{equation}
\begin{equation}
err = \underset{n}{average}(\|XYZ_0^{'} - XYZ_0\|),
\end{equation}
where $RT_{01}^{fw}$ denotes the rotation and translation computed from forward flow, which carries the pointcloud from timestep 0 to timestep 1, and $RT_{10}^{bw}$ is the backward counterpart.

If the average displacement across the entire pointcloud is above a threshold (set to 0.1 meters), then we treat the egomotion estimate for that image pair as “incorrect”, and remove that pair of frames from the analysis. We keep a minimum of 10 views and a maximum of 25 views (by taking 25 views with lowest error) from this process. 

\end{document}